\documentclass{article}

\usepackage{microtype}

\usepackage[accepted]{icml2024}
\usepackage{xcolor}
\usepackage{hyperref}
\hypersetup{
    colorlinks=true,
    linkcolor=violet,
    citecolor=violet,
    urlcolor=blue
}
\usepackage{mdframed}
\usepackage{listings}
\usepackage{caption}
\usepackage{tabularray}
\UseTblrLibrary{booktabs}
\SetTblrInner[booktabs]{abovesep=0pt, belowsep=0pt, rowsep=0.2pt}
\SetTblrInner[booktabs]{cells = {cmd=\small}}
\NewTableCommand\seprule{\specialrule{\lightrulewidth,gray8}{2pt}{2pt}}
\NewTableCommand\uniquerule{\specialrule{\lightrulewidth,gray7,dashed}{2pt}{2pt}}
\definecolor{lightb}{RGB}{235,245,255}

\usepackage{graphicx}

\usepackage{pifont}%

\usepackage[nointegrals]{wasysym}
\usepackage{enumitem} %
\setlist[itemize]{leftmargin=*}

\usepackage{xspace}
\usepackage{xcolor}
\usepackage{mathtools}

\usepackage[capitalise, noabbrev]{cleveref}
\crefformat{section}{\S#2#1#3}
\Crefformat{section}{\S#2#1#3}

\usepackage{fontawesome}

\newcommand\COMMENT[1]{}
\newcommand\Comment[1]{}

\usepackage[normalem]{ulem}
\definecolor{applegreen}{rgb}{0.55, 0.71, 0.0}

\newcommand\passat[1]{\mbox{pass@{#1}}}
\newcommand\Passat[1]{\mbox{Pass@{#1}}}

\newcommand\tech{\textsc{OSS-Instruct}\xspace}
\newcommand\codellamaPyAbbrev{CL}
\newcommand\starcoderAbbrev{SC}
\newcommand\dscoderAbbrev{DS}
\newcommand\model{\mbox{Magicoder}\xspace}
\newcommand\modelc{Magicoder-\codellamaPyAbbrev{}\xspace}
\newcommand\modeld{Magicoder-\dscoderAbbrev{}\xspace}
\newcommand\modelx{\mbox{Magicoder$\mathcal{S}$}\xspace}
\newcommand\modelxd{\modelx{}-\dscoderAbbrev\xspace}
\newcommand\modelxc{\modelx{}-\codellamaPyAbbrev\xspace}
\newcommand\evolinstruct{Evol-Instruct\xspace}
\newcommand\evolcode{\texttt{evol-codealpaca-v1}\xspace}
\newcommand\selfinstruct{\textsc{Self-Instruct}\xspace}
\newcommand\octopack{OctoPack}
\newcommand\llm{LLM\xspace}
\newcommand\llmfull{Large Language Model\xspace}
\newcommand\github{GitHub\xspace}
\newcommand\codealpaca{Code Alpaca\xspace}
\newcommand\codesearchnet{CodeSearchNet\xspace}

\newcommand\swebench{SWE-bench\xspace}
\newcommand\repoeval{RepoEval\xspace}
\newcommand\crosscodeeval{\textsc{CrossCodeEval}\xspace}
\newcommand\seedcorpus{\texttt{starcoderdata}\xspace}
\newcommand\thestack{The Stack\xspace}
\newcommand\hugging{Hugging~Face\xspace}

\newcommand\adafactor{Adafactor\xspace}
\newcommand\bigcodeharness{\texttt{bigcode-evaluation-harness}}
\newcommand\tfidf{TF-IDF\xspace}

\newcommand\instructor{INSTRUCTOR\xspace}
\newcommand\nsamples{\texttt{num\_samples}}
\newcommand\temperature{\texttt{temperature}}
\newcommand\topp{\texttt{top\_p}}
\newcommand\maxLen{\texttt{max\_length}}

\newcommand{\eg}{\emph{e.g.,}\xspace}
\newcommand{\ie}{\emph{i.e.,}\xspace}

\newcommand\llama{\textsc{Llama}}
\newcommand\codellamapy{\textsc{\codellamaTerm\codellamaSpacing{}\codellamaHyphen{}Python}}
\newcommand\codellamainst{\textsc{\codellamaTerm\codellamaHyphen{}Instruct}}
    \newcommand\codellamaSpacing{}
    \newcommand\codellamaTerm{Code\codellamaSpacing{}Llama}
    \newcommand\codellamaHyphen{\codellamaSpacing{}-\codellamaSpacing{}}
    \newcommand\codellama{\textsc{\codellamaTerm}}
\newcommand{\starcoderbase}{StarCoderBase}
\newcommand{\starcoder}{StarCoder}
\newcommand{\wizardcoder}{WizardCoder}
\newcommand{\wizardcoderc}{WizardCoder-\codellamaPyAbbrev}
\newcommand{\wizardcodersc}{WizardCoder-\starcoderAbbrev}

\newcommand\chatgpt{ChatGPT}
\newcommand\gptthreefive{GPT-3.5}
\newcommand\gptthreefiveturb{\gptthreefive~Turbo}
\newcommand\gptfour{GPT-4}
\newcommand\gptfourturb{GPT-4~Turbo}
\newcommand{\mixtralfull}{Mixtral-8x7B-Instruct-v0.1}
\newcommand{\mixtral}{Mixtral}

\newcommand\codetf{CodeT5}

\newcommand\incoder{\textsc{InCoder}}
\newcommand\codegen{CodeGen}

\newcommand\gptthree{GPT-3}
\newcommand\alpaca{Alpaca}

\newcommand\wizardlm{WizardLM}

\newcommand\dscoder{DeepSeek-Coder}
\newcommand\dscoderbase{DeepSeek-Coder-Base}
\newcommand\dscoderinst{DeepSeek-Coder-Instruct}
\newcommand\codegenmono{CodeGen-Mono}
\newcommand\codecush{Code-Cushman-001}
\newcommand\mistral{Mistral}

\newcommand\datagenmodel{\texttt{gpt-3.5-turbo-1106}}

\newcommand\humaneval{HumanEval}
\newcommand\humanevalp{HumanEval+}
\newcommand\evalplus{EvalPlus}
\newcommand\mbpp{MBPP}
\newcommand\mbppp{MBPP+}
\newcommand\apps{APPS}
\newcommand\codecontests{CodeContests}
\newcommand\dsonek{DS-1000}
    \newcommand\tensorflow{TensorFlow}
    \newcommand\torch{PyTorch}
    \newcommand\matplotlib{Matplotlib}
    \newcommand\numpy{NumPy}
    \newcommand\scipy{SciPy}
    \newcommand\sklearn{Sklearn}
    \newcommand\pandas{Pandas}
\newcommand\gsm{GSM8K}
\newcommand\multiple{MultiPL-E}
\newcommand\python{Python}
\newcommand\js{JavaScript}
\newcommand\java{Java}
\newcommand\cxx{C++}
\newcommand\swift{Swift}
\newcommand\php{PHP}
\newcommand\rust{Rust}
\renewcommand\cite{\citep}	%

\usepackage[commandnameprefix=always]{changes}
\setanonymousname{Yuxiang}
\setauthormarkuptext{name}

\icmltitlerunning{\model: Empowering Code Generation with \tech{}}

\begin{document}

\twocolumn[
\icmltitle{\raisebox{-1.7mm}{\includegraphics[width=7mm]{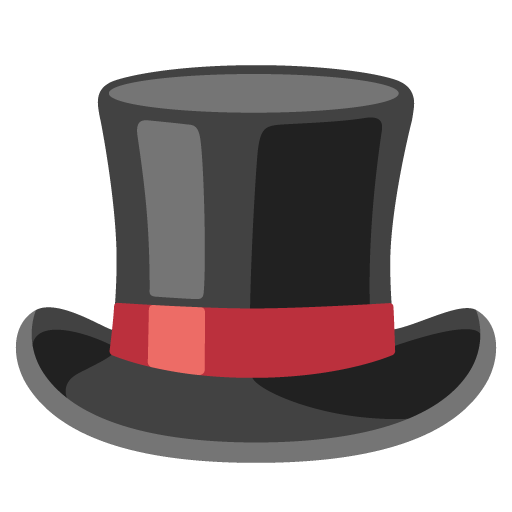}}\hspace{1pt}\model: Empowering Code Generation with \tech{}}

\icmlsetsymbol{equal}{*}
\icmlsetsymbol{intern}{$\dagger$}

\begin{icmlauthorlist}
\icmlauthor{Yuxiang Wei}{uiuc}
\icmlauthor{Zhe Wang}{thu,intern}
\icmlauthor{Jiawei Liu}{uiuc}
\icmlauthor{Yifeng Ding}{uiuc}
\icmlauthor{Lingming Zhang}{uiuc}
\end{icmlauthorlist}

\icmlaffiliation{uiuc}{University of Illinois at Urbana-Champaign, USA}
\icmlaffiliation{thu}{Tsinghua University, China}

\icmlcorrespondingauthor{Yuxiang Wei}{ywei40@illinois.edu}

\icmlkeywords{Large Language Models, Code Generation, Instruction Tuning}

\vskip 0.3in
]

\printAffiliationsAndNotice{\textsuperscript{$\dagger$}The work was done during a remote summer internship at the University of Illinois.}

\begin{abstract}
We introduce \model{}, a series of \emph{fully open-source} (code, weights, and data) \llmfull{s} (\llm{s}) for code 
that significantly closes the gap with top code models while having no more than 7B parameters. 
\model{} models are trained on 75K synthetic instruction data using \tech,
a novel approach to enlightening \llm{s} with open-source code snippets to generate diverse instruction data for code.
Our main motivation is to mitigate the inherent bias of
{the synthetic data generated by \llm{s}} through the wealth of open-source references for the production of more realistic and controllable data.
The orthogonality of \tech{} and other data generation methods like \evolinstruct{} further enables us to build an enhanced \modelx{}.
Both \model{} and \modelx{} substantially outperform state-of-the-art code models with similar or even larger sizes on a wide range of coding benchmarks.
Notably, \modelxc-7B based on \codellama{} even surpasses the
{prominent} \chatgpt{} on
{\humanevalp{}}
(66.5 vs. 65.9 in \passat{1}).
Overall, \tech{} opens a new direction for
crafting diverse synthetic instruction data for code using abundant open-source references.

\end{abstract}

\begin{figure*}[ht]
\centering
\includegraphics[width=0.9\linewidth]{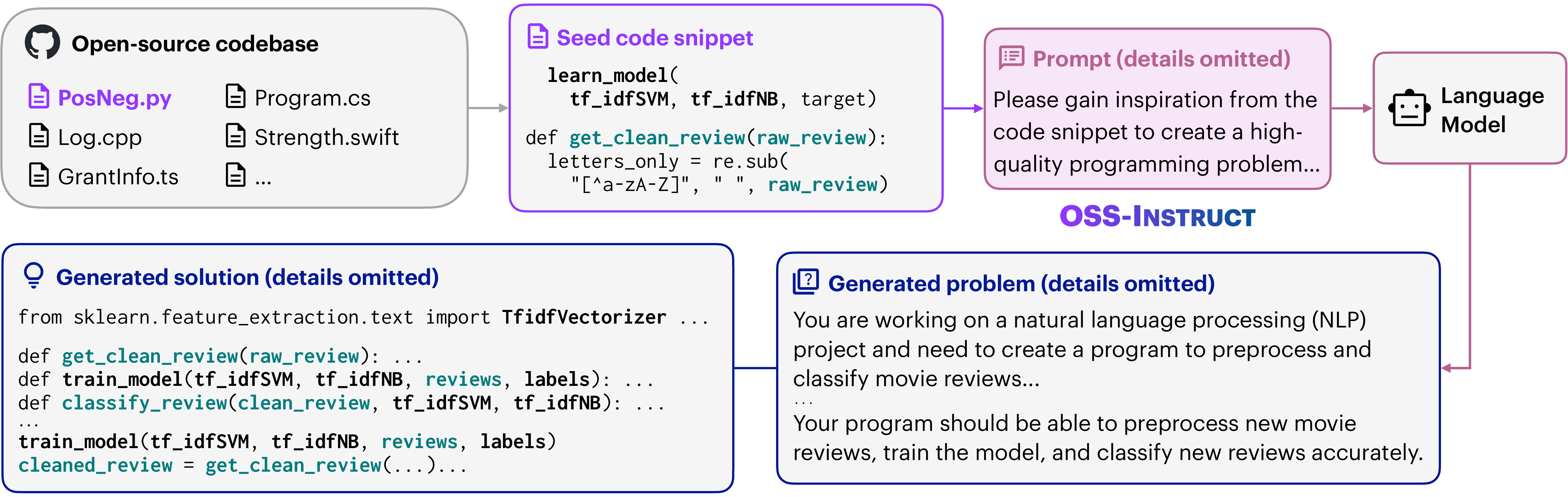}\\[8pt]%
\includegraphics[width=0.9\linewidth]{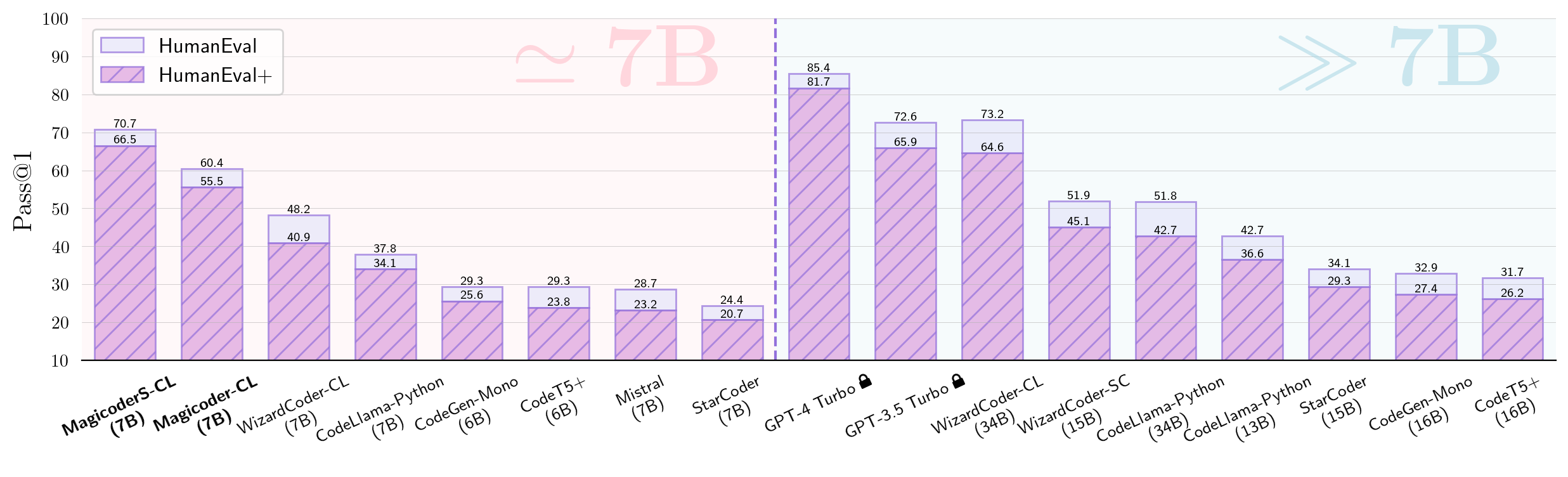}
\caption{Overview of \tech{} and the \passat{1} results of different \llm{s} on \humaneval~(+)}
\label{fig:overview}
\end{figure*}

\section{Introduction}
{
Code generation, also known as program synthesis~\cite{PGL-010}, is a long-standing challenge in computer science.
In the past few decades, a large body of research has been studying symbolic approaches, such as abstraction-based synthesis~\cite{abstractRefinement, conflictDriven} for general-purpose synthesis problems and programming by examples~\cite{flashfill, liu2023neuri} for domain-specific tasks. %
Until recently, \llmfull{s} (\llm{s}) trained on code~\cite{synthesisllm,codex} has shown outstanding breakthroughs in generating code that accurately satisfies user intents, and they are widely deployed to assist real-world software development~\cite{copilot,codewhisperer}.
}

Initially, closed-source models such as \gptthreefiveturb~\cite{chatgpt}~(\ie \chatgpt) and \gptfour~\cite{gpt4} massively dominated various coding benchmarks and leaderboards~\cite{codex,synthesisllm,liu2023code,ds1000, xia2023keep}.
To further push the boundaries of code generation with open source \llm{s}, \emph{\selfinstruct{}}~\cite{selfinstruct} is adopted to bootstrap
{the instruction-following ability of \llm{s}}.
In the realm of code, practitioners commonly devise synthetic coding instructions using a stronger teacher model (\eg \chatgpt{} and \gptfour) and then finetune a weaker student model (\eg \codellama~\cite{codellama}) with the generated data to distill the knowledge from the teacher~\cite{alpaca, codealpaca}.
For example, \codealpaca{}~\cite{codealpaca} consists of 20K automatically generated code instructions by applying \selfinstruct{} on \chatgpt{} using 21 seed tasks.
To further enhance the coding abilities of \llm{s}, \citet{wizardcoder} proposes \emph{Code \evolinstruct} that employs various heuristics to increase the complexity of seed code instructions (\codealpaca{} in this case), achieving state-of-the-art (SOTA) results among open-source models.

While these data generation methods can effectively improve the instruction-following capability of an \llm, they rely on a narrow range of predefined tasks or heuristics under the hood.
For example, on the one hand, \codealpaca{} that adopts \selfinstruct{} only relies on \emph{21 seed tasks} to generate new code instructions using an identical prompt template.
On the other hand, Code \evolinstruct{} takes \codealpaca{} as seeds and merely depends on \emph{5 heuristics} to evolve the dataset.
As partly suggested by \citet{attrprompt} and \citet{selfinstruct}, such approaches may significantly inherit the system bias inherent in the \llm{s} as well as the predefined tasks.

Therefore, in this paper, we propose \tech{} to mitigate the inherent bias of \llm{s} and to unleash their potential to craft diverse and creative code instructions via direct learning from the open source.
As shown in \Cref{fig:overview}, \tech{} leverages a powerful \llm{} to automatically generate new coding problems by \emph{drawing inspiration} from any random code snippets collected from the open source.
In this example, the \llm{} gets inspired by two incomplete code fragments from different functions and manages to relate them and craft a realistic machine learning problem.
Thanks to the ``infinite'' real-world open-source code, \tech{} can directly produce \emph{diverse}, \emph{realistic}, and \emph{controllable} code instructions by providing distinct seed code snippets.
In the end, we generate 75K synthetic data to finetune \codellamapy{}-7B, resulting in \modelc{}.
While being simple and effective, \tech{} is orthogonal to existing data generation methods, and they can be combined to further boost the models' coding capabilities. 
Therefore, we continually finetune \modelc{} on an open-source \evolinstruct{} dataset with 110K entries, producing \modelxc{}.

We evaluate \model{} and \modelx{} on a wide range of coding tasks, including \humaneval~\cite{codex} and \mbpp~\cite{synthesisllm} for \python{} text-to-code generation,
\multiple~\cite{multiple} for multilingual code completion,
and \dsonek~\cite{ds1000} for solving data science problems.
We further adopt \evalplus~\cite{liu2023code}, which includes the augmented \humanevalp{} and \mbppp{} datasets for more rigorous model evaluation. 
Both \modelc{} and \modelxc{} substantially boost the base \codellamapy-7B. Additionally, \modelc{} even outperforms \wizardcoderc{-7B}, \wizardcodersc{-15B}, and all studied SOTA \llm{s} with less than or equal to 16B parameters on all the benchmarks we tested.
Also, the \passat{1} result of the enhanced \modelxc{} is on par with \chatgpt{} on \humaneval{} (70.7 vs. 72.6) and surpasses it on the more rigorous \humanevalp{} (66.5 vs. 65.9), indicating that \modelxc{} can generate more robust code.
It also achieves SOTA results among all code models at the same scale.

Additionally, we notice a very recent advancement in the development of the \dscoder{} series~\cite{dscoder} which has shown exceptional coding performance.
However, due to the limited technical details disclosed, we only briefly discuss them in \Cref{sec:comparison-with-dscoder}.
Despite this, we applied \tech{} on \dscoderbase~6.7B, resulting in the creation of \modeld{} and \modelxd{}.
In addition to the consistent findings on the previous results with \codellamapy{}-7B as the base model,
\modeld{} and \modelxd{} benefit from the more powerful \dscoderbase-6.7B.
This advantage is demonstrated by \modelxd, which achieves a remarkable 76.8 \passat{1} on \humaneval.
\modelxd{} also outperforms \dscoderinst-6.7B on \humaneval~(+) and \mbpp~(+) with 8$\times$ less finetuning tokens.

To justify the design of \tech{}, \ie generating instruction-tuning data from open-source references rather than using the references directly, we demonstrate that finetuning the base models with semantically relevant comment-function pairs extracted from open-source projects even negatively impacts the model performance (\Cref{sec:direct-tune}).

In general, we make the following contributions:
\begin{itemize}

\item We introduce \tech, a pioneering approach to enlightening
\llm{s} with open-source code snippets to generate more diverse, realistic, and controllable coding instruction data, which can be leveraged to substantially boost the performance of various \llm{s} via instruction tuning.
It opens a new dimension for creating low-bias and diverse instruction-tuning data from the abundance of open-source references.

\item We build the \model{} series trained with \tech{} and \modelx{} series trained on a combination of \tech{} and \evolinstruct{}.
Our evaluation across 6 benchmarks shows that all \model{s} significantly improve the base \llm{s}.
Notably, both \modelxc{} and \modelxd{} outperform \chatgpt{} on \humanevalp{} with only \emph{7B parameters}.

\item We fully open source the model weights, training data, and source code 
at \url{https://github.com/ise-uiuc/magicoder}
to facilitate future research.

\end{itemize}

\section{\tech{}: Instruction Tuning from Open Source}
In this section, we elaborate on our \tech{} approach.
From a high level, as shown in \Cref{fig:overview}, \tech{} works by prompting an \llm{} (\eg \chatgpt) to generate a coding problem and its solution according to some seed code snippet collected from the wild (\eg from \github).
The seed snippet offers controllability of the generation and encourages the \llm{} to create diverse coding problems that can reflect real-world programming scenarios.

\subsection{Generating Coding Problems}

\tech{} is powered by seed code snippets that can be easily collected from open source.
In this work, we directly adopt \seedcorpus{} as our seed corpus, a filtered version of \thestack{}~\cite{thestack} dataset that \starcoder{} is trained on, containing permissively licensed source code documents in various programming languages.
We chose \seedcorpus{} because it is widely adopted, includes massive high-quality code snippets, and is even post-processed for data decontamination~\cite{starcoder, santacoder}.
For each code document from the corpus, we randomly extract 1--15 consecutive lines as the seed snippet for the model to gain inspiration from and produce coding problems. In total, we collected 80K initial seed snippets from 80K code documents, 40K from Python, and 5K from each of C++, Java, TypeScript, Shell, C\#, Rust, PHP, and Swift respectively.
Then, each collected seed code snippet is applied to the prompt template shown in \Cref{apd:prompt}, which a teacher model takes as input and outputs both a coding problem and its solution.

\subsection{Data Cleaning and Decontamination}
\label{sec:tech:decontaminate}

We perform data cleaning by excluding samples that are identical or share the same seed code snippet.
While there exist other sorts of noisiness (\eg the solution is incomplete)
in the generated data, inspired by \citet{unnatural}, they are not removed as we believe they still contain valuable information for \llm{s} to learn.
More experimental details can be found in \Cref{apd:noise}.
Finally, we apply the same logic as \starcoder~\citet{starcoder} to decontaminate our training data by removing coding problems that contain docstrings or solutions from \humaneval{}~\cite{codex} and \mbpp{}~\cite{synthesisllm}, docstrings from \apps{}~\cite{apps}, prompts from \dsonek{}~\cite{ds1000}, or questions from \gsm{}~\cite{gsm8k}.
As part of our analysis, the decontamination procedure only filters out 9 additional samples. Since the seed corpus \seedcorpus{} has already gone through rigorous data decontamination, this observation suggests that \tech{} is unlikely to introduce additional data leakage beyond the seeds.
The eventual \tech{} dataset contains about 75K entries.
{An overview of the dataset statistics can be found in \Cref{apd:dataset}.}

\subsection{Qualitative Examples of \tech{}}
\begin{figure*}
\centering
\includegraphics[width=0.97\textwidth]{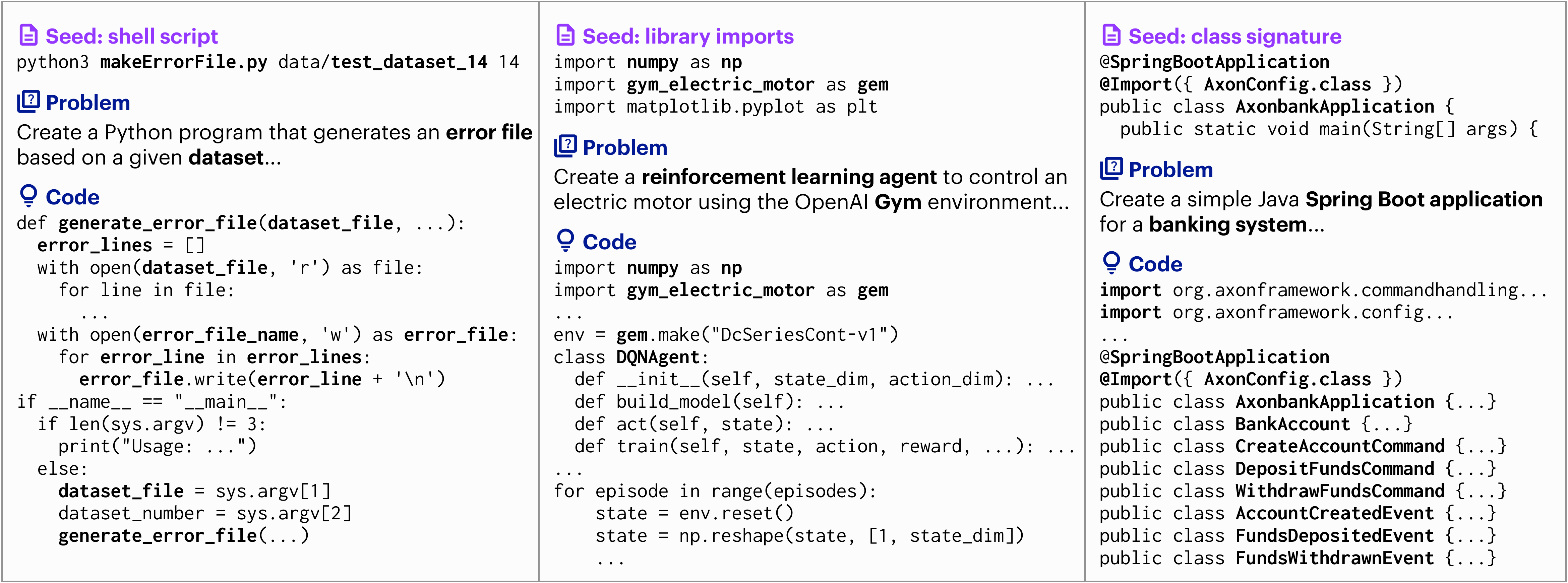}
\caption{Examples showing how \tech{} generates problems and solutions from seed code snippets. Detailed problem requirements, implementations, and explanations are omitted for brevity. More examples can be found in \Cref{apd:examples}.}
\label{fig:examples}
\end{figure*}

\Cref{fig:examples}
shows some qualitative examples of how \tech{} can help \llm{} get inspiration from a seed code snippet to create new coding problems and solutions.
For example, the \emph{shell script} example shows how an \llm{} crafts a \python{} coding problem with just one line of shell script.
The \emph{library imports} example demonstrates how an \llm{} can create a realistic machine learning problem using just a few import statements. Meanwhile, the \emph{class signature} instance illustrates the ability of \llm{} to draw inspiration from an incomplete class definition featuring annotations like \texttt{\@SpringBootApplication} and keywords such as \texttt{bank}. From this, the \llm{} generates a problem that requires implementing a complete banking system based on Spring Boot.
Overall, \tech{} can inspire an \llm{} with distinct code structures and semantics to create diverse coding tasks, including algorithmic challenges, realistic issues, single-function code generation, library-based program completion, whole-program development, and even whole-application construction.

\begin{figure}
\centering
\includegraphics[width=\linewidth]{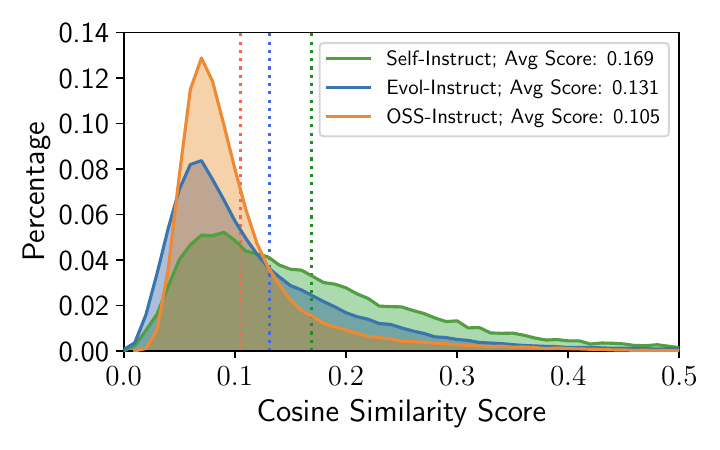}
\caption{Cosine similarities between \humaneval{} and synthetic data generated by different methods.}
\label{fig:humaneval-sim}
\end{figure}

\paragraph{Similarity with \humaneval{}}
\label{sec:humaneval-sim}
To study whether our data generation process produces more \humaneval-like problems or solutions that contribute to high performance,
we pair each sample from our 75K dataset with each of the 164 \humaneval{}~\cite{codex} samples and compute their cosine similarity using \tfidf{}~\cite{tfidf} embeddings.
We then associate each \tech{} sample with a \humaneval{} sample with the highest similarity score.
We also compare our dataset against \codealpaca, a 20K dataset applying \selfinstruct{} to code,
and \evolcode{}~\cite{evolcode}, an open-source reproduction of \evolinstruct{} containing 110K coding instructions.
We resort to the open-source implementation because the official \emph{Code \evolinstruct{}}~\cite{wizardcoder} dataset is not released.
We decontaminate all the datasets beforehand using the same way discussed in
\Cref{sec:tech:decontaminate}.
\Cref{fig:humaneval-sim} shows that \tech{} exhibits the lowest average similarity among all the studied data generation techniques while \selfinstruct{} shows the highest average similarity.
This result indicates that the improvements from \tech{} are not merely due to including data from the same distribution.

\section{Evaluation}
We choose \codellamapy-7B and \dscoderbase~6.7B as the base \llm{s}.
To derive \model{} series, we first finetune them on 75K synthetic data generated through \tech{}.
We then obtain \modelx{} by continuing finetuning \model{} with the \evolcode{} dataset, an open-source \evolinstruct{} implementation containing about 110K samples.
{More implementation details and additional evaluation results are listed in \Cref{apd:setup,apd:evalmore}. We also present interesting use cases that reflect the effectiveness of instruction tuning in \Cref{apd:case} and demonstrate \model{}'s capability to generate complex programs in \Cref{apd:complex}.}

\subsection{Python Text-to-Code Generation}

\begin{table*}[t]
 \caption{\Passat{1} (\%) results of different \llm{s} on \humaneval{}~(+) and \mbpp{}~(+) computed with greedy decoding. The abbreviations ``CL'' and ``SC'' refer to the base models \codellamapy{} and \starcoder{}, respectively.
 We report the results consistently from the \evalplus~\cite{liu2023code} Leaderboard.
 }
\label{tab:python-text2code}
\centering
        \begin{booktabs}{
            colspec={@{}lrrrrcc@{}},
            cell{1}{1-3} = {r=2}{m},
            cell{1}{4,6} = {c=2}{c},
            column{4} = {rightsep=10pt},
            column{5} = {leftsep=10pt},
            row{Y-Z} = {bg=lightb}
        }
        \toprule
        Model & Release Date & Size & Benchmark & & Open-Source & \\
        \cmidrule[lr]{4-7}
        & & & \humaneval~(+) & \mbpp~(+) & Weight & Data \\
        \midrule
        \gptthreefiveturb & Nov 2023 & - & 72.6~~(65.9) & 81.7~~(69.4) &\Circle & \Circle\\
        \gptfourturb & Nov 2023 & - & \textbf{85.4}~~(\textbf{81.7}) & \textbf{83.0}~~(\textbf{70.7}) & \Circle & \Circle\\
        \seprule
        \codellamapy & Aug 2023 & 34B & 51.8~~(42.7) & 67.2~~(52.9) & \CIRCLE & \Circle\\
        \wizardcoderc & Sep 2023 & 34B & 73.2~~(64.6)  & 73.2~~(59.9) & \CIRCLE & \Circle \\
        \seprule
        \codetf+ & May 2023 & 16B & 31.7~~(26.2) & 54.6~~(44.4) & \CIRCLE & \CIRCLE\\
        \codegenmono & Mar 2022 & 16B & 32.9~~(27.4) & 52.6~~(43.6) & \CIRCLE & \CIRCLE\\
        \starcoder & May 2023 & 15B & 34.1~~(29.3) & 55.1~~(46.1) & \CIRCLE & \CIRCLE\\
        \codellamapy & Aug 2023 & 13B & 42.7~~(36.6) & 61.2~~(50.9) & \CIRCLE & \Circle\\
        \wizardcodersc & Sep 2023 & 15B & 51.9~~(45.1)  & 61.9~~(50.6) & \CIRCLE & \Circle \\
        \seprule
        \starcoder & May 2023 & 7B & 24.4~~(20.7) & 33.1~~(28.8) & \CIRCLE & \CIRCLE\\
        \mistral & Oct 2023 & 7B & 28.7~~(23.2) & 50.1~~(40.9) & \CIRCLE & \Circle\\
        \codetf+ & May 2023 & 6B & 29.3~~(23.8) & 51.9~~(40.9) & \CIRCLE & \CIRCLE\\
        \codegenmono & Mar 2022 & 6B & 29.3~~(25.6) & 49.9~~(42.1) & \CIRCLE & \CIRCLE\\
        \codellamapy & Aug 2023 & 7B & 37.8~~(34.1) & 57.6~~(45.4) & \CIRCLE & \Circle\\
        \wizardcoderc & Sep 2023 & 7B & 48.2~~(40.9) & 56.6~~(47.1) & \CIRCLE & \Circle \\
        \seprule
        \modelc{} & Dec 2023  & 7B & 60.4~~(55.5) & 64.2~~(52.6) & \CIRCLE & \CIRCLE \\
        \modelxc{} & Dec 2023 & 7B & \textbf{70.7}~~(\textbf{66.5}) & \textbf{68.4}~~(\textbf{56.6}) & \CIRCLE & \CIRCLE\\
        \bottomrule
\end{booktabs}%
\end{table*}

\humaneval{}~\cite{codex} and \mbpp{}~\cite{synthesisllm} are two of the most widely used benchmarks for code generation.
Each task in these benchmarks includes a task description (\eg docstring) as the prompt, where \llm{s} generate corresponding code whose correctness is checked by a handful of test cases.
Because tests in these benchmarks can be insufficient, for more rigorous evaluation, we use \humaneval{+} and \mbpp{+}, both powered by the \evalplus{} framework~\cite{liu2023code} to obtain 80$\times$/35$\times$ more tests.
Following prior work~\cite{liu2023code,selfdebug}, for each task and \llm{} we use greedy decoding to generate one sample and focus on comparing the \passat{1} metric.

We consider a wide range of baseline models, including \codellamapy~\cite{codellama}, \wizardcoder~\cite{wizardcoder}, \gptthreefiveturb~\cite{chatgpt}, \gptfourturb~\cite{gpt4}, \starcoder~\cite{starcoder}, \codetf+~\cite{codetfp}, \codegenmono~\cite{codegen}, and \mistral~\cite{mistral}.
All the results are consistently reported from the \evalplus{}~\cite{liu2023code} leaderboard (\evalplus{} hash: \texttt{1895d2f}).

\Cref{tab:python-text2code} shows the \passat{1} results of different \llm{s} on these benchmarks.
From the results, we can first observe that \modelc{} has a clear improvement over the base \codellamapy-7B, and outperforms all studied open-source models except \codellamapy-34B and \wizardcoderc-34B. Notably, \modelc{} surpasses \wizardcodersc{}-15B and has a substantial improvement on \humaneval{} and \humanevalp{} over \codellamapy-34B.
\modelxc{} demonstrates further improvements by being trained with the orthogonal \evolinstruct{} method. \modelxc{} outperforms \chatgpt{} and all other open-source models on \humanevalp{}. Moreover, although it scores slightly lower than \wizardcoderc-34B and \chatgpt{} on \humaneval{}, it surpasses both of them on the more rigorous \humanevalp{} dataset, indicating that \modelxc{} may produce more robust code.

\subsection{Multilingual Code Generation}

In addition to \python, as shown in \Cref{tab:multilang}, we perform an extensive evaluation on 6 widely used programming languages, \ie \java, \js, \cxx, \php, \swift, and \rust, using the \multiple{} benchmark~\cite{multiple}.
We report available results from the \wizardcoder{} paper~\cite{wizardcoder} and evaluate our models consistently through \bigcodeharness~\cite{bigcode-evaluation-harness}.
We skip proprietary models such as \chatgpt{} and \gptfour{} as they are not supported by the framework.
Due to a significant inference latency when running \wizardcoderc-7B using the harness in our environment, we choose not to include it in our analysis.

The results indicate that \modelc{} improves the base \codellamapy{-7B} by a large margin among all the studied programming languages.
Moreover, \modelc{} also achieves better results than the SOTA 15B \wizardcodersc{} among half of the programming languages.
Additionally, \modelxc{} demonstrates further improvement over \modelc{} on all programming languages, achieving comparable performance against \wizardcoderc-34B with only 7B parameters.
It is worth noting that \modelc{} is only trained with very limited multilingual data but still outperforms other \llm{s} with similar or even larger sizes.
Also, although the harness evaluates models in \emph{completion} formats which are for base models, \model{s} still show significant improvements despite being only \emph{instruction-tuned}. This implies that \llm{s} can learn knowledge from the data beyond its format.

\begin{table*}
\caption{\Passat{1} results of different \llm{s} on \multiple~\cite{multiple} following the same hyperparameter settings as the \wizardcoder{} paper~\cite{wizardcoder}: $\temperature=0.2$, $\topp=0.95$, $\maxLen=512$, and $\nsamples=50$.
We evaluate all 7B models using \bigcodeharness{}~\cite{bigcode-evaluation-harness} and report other results from \wizardcoder{}.}
\label{tab:multilang}
\centering
\begin{booktabs}{
    colspec={@{}lrrrrrrr@{}},
    cell{1}{3}={c=6}{c},
    cell{1}{1-2}={r=2}{m},
    cell{6}{3-Z}={font=\bfseries},
    cell{Z}{3-Z}={font=\bfseries},
    row{Y-Z} = {bg=lightb}
}
\toprule
Model & Size & Programming Language & & & & & \\
\cmidrule[lr]{3-8}
& & \java & \js & \cxx & \php & \swift & \rust \\
\midrule

\codellama & 34B & 40.2 & 41.7 & 41.4 & 40.4 & 35.3 & 38.7 \\
\codellamapy & 34B & 39.5 & 44.7 & 39.1 & 39.8 & 34.3 & 39.7 \\
\codellamainst & 34B & 41.5 & 45.9 & 41.5 & 37.0 & 37.6 & 39.3 \\
\wizardcoderc & 34B & 44.9 & 55.3 & 47.2 & 47.2 & 44.3 & 46.2 \\

\seprule

\starcoderbase & 15B & 28.5 & 31.7 & 30.6 & 26.8 & 16.7 & 24.5 \\
\starcoder & 15B & 30.2 & 30.8 & 31.6 & 26.1 & 22.7 & 21.8 \\
\wizardcodersc & 15B & 35.8 & 41.9 & 39.0 & 39.3 & 33.7 & 27.1 \\

\seprule

\codellama & 7B & 29.3 & 31.7 & 27.0 & 25.1 & 25.6 & 25.5 \\
\codellamapy & 7B & 29.1 & 35.7 & 30.2 & 29.0 & 27.1 & 27.0 \\
\seprule
\modelc & 7B & 36.4 & 45.9 & 36.5 & 39.5 & 33.4 & 30.6 \\
\modelxc & 7B & 42.9 & 57.5 & 44.4 & 47.6 & 44.1 & 40.3 \\
\bottomrule
\end{booktabs}
\end{table*}

\begin{table*}
\centering
\caption{\Passat{1} results on \dsonek{} (completion format) with $\temperature=0.2$, $\topp=0.5$, $\maxLen=1024$, and $\nsamples=40$, following the same hyperparameter setting used in \wizardcoder~\cite{wizardcoder}.
We evaluate all the 7B models with their preferred prompt formats and report other results from \wizardcoder{}.
}
\label{tab:ds1000}
\newcommand\rotateHeader[1]{\rotatebox{0}{#1}}
 \begin{booktabs}{
        columns = {colsep=4pt},
        colspec={@{}lrrrrrrrrr@{}},
        cell{1}{1-2} = {valign=f},
        cells = {cmd={}},
        row{Y-Z} = {bg=lightb},
        cell{Z}{3-Z}={font=\bfseries}
    }       
        \toprule
        Model & Size & {+ 155\\\matplotlib} & {+ 220\\\numpy} & {+ 291\\\pandas} & {+ 68\\\torch} & {+ 106\\\scipy} & {+ 115\\\sklearn} & {+ 45\\\tensorflow} & {= 1000\\Overall}\\
        \midrule
        \incoder & 6.7B & 28.3 & 4.4 & 3.1 & 4.4 & 2.8 & 2.8 & 3.8 & 7.4\\
        \codegenmono & 16B & 31.7 & 10.9 & 3.4 & 7.0 & 9.0 & 10.8 & 15.2 & 11.7\\
        \codecush & - & 40.7 & 21.8 & 7.9 & 12.4 & 11.3 & 18.0 & 12.2 & 18.1\\
        \seprule
        \starcoder & 15B & 51.7 & 29.7 & 11.4 & 21.4 & 20.2 & 29.5 & 24.5 & 26.0\\
        \wizardcodersc & 15B & 55.2 & 33.6 & 16.7 & 26.2 & 24.2 & 24.9 & 26.7 & 29.2\\
        \seprule
        \codellamapy & 7B & 55.3 & 34.5 & 16.4 & 19.9 & 22.3 & 17.6 & 28.5 & 28.0 \\
        \wizardcoderc & 7B & 53.5 & 34.4 & 15.2 & 25.7 & 21.0 & 24.5 & 28.9 & 28.4 \\
        \seprule
        \modelc & 7B & 54.6 & 34.8 & 19.0 & 24.7 & 25.0 & 22.6 & 28.9 & 29.9\\
        \modelxc & 7B & 55.9 & 40.6 & 28.4 & 40.4 & 28.8 & 35.8 & 37.6 & 37.5\\
        \bottomrule
    \end{booktabs}%
\end{table*}
\begin{table*}
 \caption{\Passat{1} (greedy decoding) comparison between \model{} and \dscoder{}~\cite{dscoder} on \humaneval{}~(+) and \mbpp{}~(+). \dscoder{} results are reported from \evalplus{}~\cite{liu2023code} Leaderboard.}
        \centering
\label{tab:dscoder-text2code}
        \begin{booktabs}{
            colspec={@{}lrrrrcc@{}},
            cell{1}{1-3} = {r=2}{m},
            cell{1}{4,6} = {c=2}{c},
            column{4} = {rightsep=10pt},
            column{5} = {leftsep=10pt},
            row{Y-Z} = {bg=lightb}
        }
        \toprule
        Model & Size & {Training Tokens} & Benchmark & & Open-Source &\\
        \cmidrule[lr]{4-7}
        & & & \humaneval~(+) & \mbpp~(+) & Weight & Data \\
        \midrule
        \SetCell[r=3]{m}{\dscoderbase}
            & 1.3B & 2T & -  & 55.4~~(46.9) & \CIRCLE & \Circle \\
            & 6.7B & 2T & 47.6~~(39.6)  & 70.2~~(56.6) & \CIRCLE & \Circle \\
            & 33B & 2T & 51.2~~(43.3)  & - & \CIRCLE & \Circle \\
        \seprule
        \SetCell[r=3]{m}{\dscoder{} Instruct}
            & 1.3B & +2B & 64.6~~(58.5)  & 63.7~~(53.1) & \CIRCLE & \Circle \\
            & 6.7B & +2B & \textbf{73.8}~~(\textbf{70.1})  & \textbf{72.7}~~(\textbf{63.4}) & \CIRCLE & \Circle \\
            & 33B & +2B & \textbf{78.7}~~(\textbf{72.6})  & \textbf{78.7}~~(\textbf{66.7}) & \CIRCLE & \Circle \\
        \seprule
        \modeld{}  & 6.7B & +90M & 66.5~~(60.4) & 75.4~~(61.9) & \CIRCLE & \CIRCLE \\
        \modelxd{}  & 6.7B & +240M & \textbf{76.8}~~(\textbf{70.7}) & \textbf{75.7}~~(\textbf{64.4}) & \CIRCLE & \CIRCLE \\
        \bottomrule
\end{booktabs}%
\end{table*}

\subsection{Code Generation for Data Science}

The \dsonek{} dataset~\cite{ds1000} contains 1K distinct data science coding issues ranging from 7 popular data science libraries in \python.
It evaluates the realistic and practical use case of an \llm{} and offers unit tests for validating each problem.
\dsonek{} has both \emph{completion} and \emph{insertion} modes, but here we only evaluate \emph{completion} because the base \codellamapy{} does not support infilling.
\Cref{tab:ds1000} shows the evaluation results where 
we include the recent \incoder~\cite{incoder}, \codegen~\cite{codegen}, \codecush~\cite{codecush}, \starcoder~\cite{starcoder}, \codellamapy~\cite{codellama}, and \wizardcoder~\cite{wizardcoder}.
We can see from the table that \modelc-7B already outperforms all the baselines we evaluate, including state-of-the-art \wizardcoderc-7B and \wizardcodersc-15B. \modelxc{}-7B further breaks the limit by introducing an 8.3 percentage point absolute improvement over \wizardcodersc-15B.

\subsection{Comparison with \dscoder}
\label{sec:comparison-with-dscoder}

\dscoder{}~\cite{dscoder} is a series of models released concurrently to our work and they demonstrate superior coding performance.
We only briefly discuss it in this section because its data and instruction tuning details are not publicly available at the time of writing.
We apply the same finetuning strategy on \dscoderbase-6.7B as we performed on \codellamapy-7B, leading to \modeld{} and \modelxd{}.
\Cref{tab:dscoder-text2code} shows a similar trend as \Cref{tab:python-text2code} that the base model can be significantly improved after applying \tech{}.
Remarkably, the \modelxd{} variant surpasses \dscoderinst-6.7B on all the benchmarks with $\times$8 fewer training tokens, and it also closely matches \dscoderinst-33B on these datasets.

\section{Ablations of Data Source}
\subsection{Impact of the Language Distribution}

To understand the correlation between the programming languages appearing in the training data and the downstream performance of different languages, we conduct an additional ablation study about the training data.
We classify the 75K training data into approximately 43K \python-only, and 32K non-\python{} data according to whether \verb|```python| is a substring of the generated data.
We do not classify the data based on the seed code snippet because \llm{}s performing \tech{} may produce code in a different programming language than the seed.

\begin{table*}
\caption{Ablation study of using different programming languages as training data.
We show the \passat{1} results on \humanevalp{}~\cite{liu2023code} for \python{} and the average \passat{1} results on \multiple{}~\cite{multiple} for the same set of programming languages used in \Cref{tab:multilang} (\ie \java, \js, \cxx, \php, \swift, and \rust).
All the variants are finetuned with 2 epochs and evaluated through greedy-decoding.
}
\label{tab:ablation}
\centering
\begin{booktabs}{
    colspec={@{}llrr@{}},
}
\toprule
Model~(7B) & Finetuning Data & \python~(\humanevalp) & Others~(\multiple) \\
\midrule
\codellamapy & - & 34.1 & 29.6 \\
\seprule
\modelc & \python{} (43K) & 47.6 & 32.7 \\
\modelc & Others (32K) & 44.5 & \textbf{38.3} \\
\seprule
\modelc & Both (75K) & \textbf{55.5} & 37.8  \\
\bottomrule
\end{booktabs}%
\end{table*}

\Cref{tab:ablation} shows the evaluation results, where we consistently finetune the base \codellamapy-7B for 2 epochs on different data partitions using the same training hyperparameters explained in \Cref{apd:setup}. 
From the table, we can see that, as can be imagined, training on \python{} or non-\python{} data can substantially boost the performance of the base model in \python{} or non-\python{} tasks, respectively.
Interestingly, instruction tuning on different programming languages can still boost the overall coding performance that includes out-of-distribution languages.
For example, when trained on only non-\python{} data, \modelc{} still achieves a 10.4 percentage point improvement over the base model in the \python{}-only evaluation.
This implies \llm{s} can establish correlations between different programming languages and perform transfer learning of deeper code semantics.
Finally, we observe a more significant boost in \python{} evaluation when combining data from both sources, with a slight decrease in multilingual performance compared with only finetuning on multilingual data.
We attribute this decrease to the dominant amount of \python{} data (around 57\%) during instruction tuning.

\subsection{\tech{} vs. Direct Finetuning}
\begin{table}
\caption{Comparison between \tech{} and directly finetuning on comment-function pairs with \codellamapy-7B as the base model.}
\label{tab:pair-ablation}
\centering
\begin{booktabs}{
    colspec={@{}lrr@{}},
}
\toprule
Finetuning Data & \humanevalp &\multiple \\
\midrule
Base model w/o finetuning & 34.1 & 29.6 \\
Comment-function pairs (75K) & 34.1 & 24.1\\
\tech{} (75K) & \textbf{55.5} & \textbf{37.8} \\
\bottomrule
\end{booktabs}%
\end{table}

\label{sec:direct-tune}
The fact that \tech{} gets an \llm{} inspired from open-source code snippets may lead to a natural question: why not directly finetuning on these open-source code?
To answer this question, we follow \codesearchnet~\cite{codesearchnet} to mine semantically relevant \emph{comment-function} pairs from the same seed document corpus we use to construct the 75K \tech{} dataset.
We then train the model to predict the function bodies from the function signatures and comments.
We prioritize comment-function pairs that overlap with our \emph{75K seed snippets}, resulting in about 11K data points.
To align with our 75K samples, we collect the remaining 64K samples using the whole corpus of \emph{75K seed documents}.
Eventually, we have the same number of comment-function pairs with \tech{} data.

We finetune the base \codellamapy-7B for 2 epochs using the paired data, following the same training setup discussed in \Cref{apd:setup}.
From \Cref{tab:pair-ablation}, we observe that finetuning on 75K paired comment-function data even worsens the base model, while \tech{} helps to introduce a substantial boost.
We conjecture that the degradation is owing to the substantial noise and inconsistency that exists intrinsically in the data pairs, even though these paired data exhibit \emph{very similar} format as \humaneval{} or \multiple{} problems.
This further shows that data factuality, rather than the format, is essential to code instruction tuning. It also indicates the superiority of \tech{} which can translate these loosely related code fragments into semantically-consistent instruction-tuning data.

\subsection{\tech{} with A Less Powerful Teacher}

In this section, we explore the factors contributing to the effectiveness of \tech{} beyond just the distillation of the teacher model. We propose two potential key reasons.
First, since the base model is pretrained with comprehensive code data, the distillation process likely activates the model's internal capabilities, leading to improved performance in coding tasks.
Second, \tech{} uses seed code snippets to generate problem-solution pairs in one shot. These seed snippets provide valuable context, enabling the model to create better solutions than a plain teacher model lacking such seed information. These enhanced solutions can then be used to train more effective student models.
To verify these points, we conduct an additional experiment by generating a subset of 20K \tech{} data using \mixtralfull~\cite{mixtral}, a state-of-the-art, general-purpose, open-source \llm{}.

\begin{table}[htbp]
\caption{\Passat{1} on \humanevalp{} and \mbppp{} when finetuning \codellamapy{-7B} for 2 epochs on 20K \tech{} data generated by \mixtralfull{}~\cite{mixtral}.}
\label{tab:mixtral}
\centering
\begin{booktabs}{
    colspec={@{}lrr@{}},
}
\toprule
Model & \humanevalp{} & \mbppp\\
\midrule
\mixtralfull{} & 39.6 & 47.4\\
\codellamapy{-7B} & 34.1 & 45.4\\
\modelc{-Mixtral}-7B & \textbf{55.5} & \textbf{50.4}\\
\bottomrule
\end{booktabs}%
\end{table}

\Cref{tab:mixtral} indicates that \modelc{}-\mixtral{}-7B not only significantly improves over the base \codellamapy{}, but is also better than \mixtralfull{} (\ie the teacher model) across \humanevalp{} and \mbppp{}.
These results suggest that \tech{} is not simply distilling a teacher model, but also triggering the base model’s own capability and effectively leveraging the information encapsulated in seed code snippets.

\section{Related Work}

\paragraph{Foundation models for code}

Trained over billions of lines of code, \llm{s} have demonstrated outstanding performance in a wide range of software engineering tasks, including code generation~\cite{codex, synthesisllm}, program repair~\cite{alpharepair, wei2023copiloting, aprstudy, aprstudy2, bouzenia2024repairagent}, and software testing~\cite{fuzz4all, titanfuzz, yuan2023no, schafer2023empirical, lemieux2023codamosa}.
In particular, prominent base models, such as \codegen{}~\cite{codegen}, \codetf~\cite{codet5}, \starcoder{}~\cite{starcoder}, and \codellama{}~\cite{codellama}, are pre-trained over a huge number of codebase from scratch, establishing the fundamental ability of general code generation and understanding.
More recent code \llm{s}, such as \dscoder{}~\cite{dscoder} and StarCoder2~\cite{starcoder2}, additionally organize the pretraining data at the repository level to enhance the model's contextual understanding capabilities.
Furthermore, these base models are also finetuned~\cite{wizardcoder} or prompted~\cite{selfdebug} to unlock their true potential to specialize in solving domain-specific coding tasks.

\paragraph{Instruction tuning with synthetic data}
Instruction tuning aims to improve
pretrained \llm{s} by finetuning them with a mixture of instructions and corresponding responses~\cite{wei2022finetuned}.
However, obtaining high-quality instructional data
{is oftentimes laborious}.
Hence, researchers are increasingly focusing on the development of methods to generate synthetic instruction data.
\citet{selfinstruct} introduces \emph{\selfinstruct{}}, where a foundation \llm{} (\gptthree~\cite{gpt3}) is used to generate synthetic instruction-response pairs with carefully crafted prompts. The \emph{same} \llm{} is then instruction-tuned on the synthetic data to distill such self-generated knowledge.
{
This technique has been further extended to create synthetic data with different \llm{s}}.
For example, \alpaca{}~\cite{alpaca} and \codealpaca{}~\cite{codealpaca} apply \selfinstruct{} to finetune \llama{} with \chatgpt-generated instructions. 
To improve \selfinstruct{}, \wizardlm~\cite{xu2023wizardlm} and \wizardcoder~\cite{luo2023wizardcoder} propose \evolinstruct{} and Code \evolinstruct{} by guiding \chatgpt{} with heuristic prompts to make the synthetic data more complex and diverse.
More recently, \citet{phi} shows that textbook-quality synthetic data alone can help the model achieve remarkable coding and reasoning capabilities.
Orthogonal to all existing methods, our proposed \tech{} allows \llm{s} to get inspired from real-world code snippets for better controllability, quality, and creativity in coding tasks.

\paragraph{Evaluating \llm{s} for code}

Most code benchmarks evaluate \llm{s} on generating \emph{single-function} programs from natural language descriptions.
Such benchmarks include \humaneval{}~\cite{codex}, \mbpp{}~\cite{synthesisllm}, \apps{}~\cite{apps}, and \codecontests~\cite{codecontests}.
A handful of manual tests are used to assess the functional correctness of \llm{-generated} solutions.
However, insufficient tests can lead to false negatives. Consequently, the \evalplus{} framework~\cite{liu2023code} produces \humaneval{+} and \mbpp{+} by extending 80$\times$/35$\times$ more tests.
To address dataset contamination issues, researchers propose LiveCodeBench~\cite{lcb}, which compiles fresh coding problems not included in model training, and EvoEval~\cite{evoeval}, which strategically leverages LLMs to evolve existing benchmarks into new coding tasks.
Meanwhile, there are comprehensive benchmarks evaluating code generation for data science (\dsonek~\cite{ds1000}), addressing open-source issues (\swebench~\cite{swebench}), and repository-level code generation (\crosscodeeval{}~\cite{cceval} and \repoeval{}~\cite{repocoder}).

\section{Conclusion and Future Work}
We propose \tech, a novel data generation method using \llmfull{}s to generate diverse coding challenges from open-source code snippets.
This approach enables \model{}, which significantly improves the base \llm. Despite having less than 7B parameters, it
can outperform all evaluate
\llm{s} with less than or equal to 16B parameters, including the 15B \wizardcoder{}.
Combining \tech{} with \evolinstruct{} allows us to build the enhanced \modelx{} models.
They achieve remarkable results by rivaling leading models like \chatgpt{} in \humaneval{} benchmarks.
We fully open source the model weights, training data, and source code, to enable future research in \llm{s} for code.
In the near future, we will apply \tech{} to larger base models. We will also continue advancing \tech{} by generating higher-quality data with a strategically designed distribution of the seed code snippets and with more advanced teacher \llm{s} such as \gptfour. 

\section*{Acknowledgement}
We thank all the reviewers for their insightful comments and suggestions for our paper. This work was partially supported by NSF grant CCF-2131943, as well as Kwai Inc.

\section*{Impact Statement}
This work is motivated to boost large language models in terms of their code generation and understanding capabilities through instruction tuning.
The proposed \tech{} method leverages the abundance of open source to generate diverse and controllable instruction data.
We expect this idea to also foster innovative software solutions tailored to domain-specific needs, particularly in areas where real data is private and scarce, by generating extensive synthetic data.
Additionally, our method reinforces the value of community-driven content and knowledge sharing by incorporating open-source code as references.

However, it is essential to recognize the potential for misuse, such as the deliberate generation of vulnerable code that can be exploited for malicious purposes.
Ultimately, adhering to ethical guidelines is crucial to ensure the responsible use of this technique.

\bibliography{reference}
\bibliographystyle{icml2024}

\newpage
\appendix
\onecolumn
\section{More Details of \tech{}}

\subsection{Prompt Design}
\label{apd:prompt}
\begin{figure*}
\centering

\includegraphics[width=\textwidth]{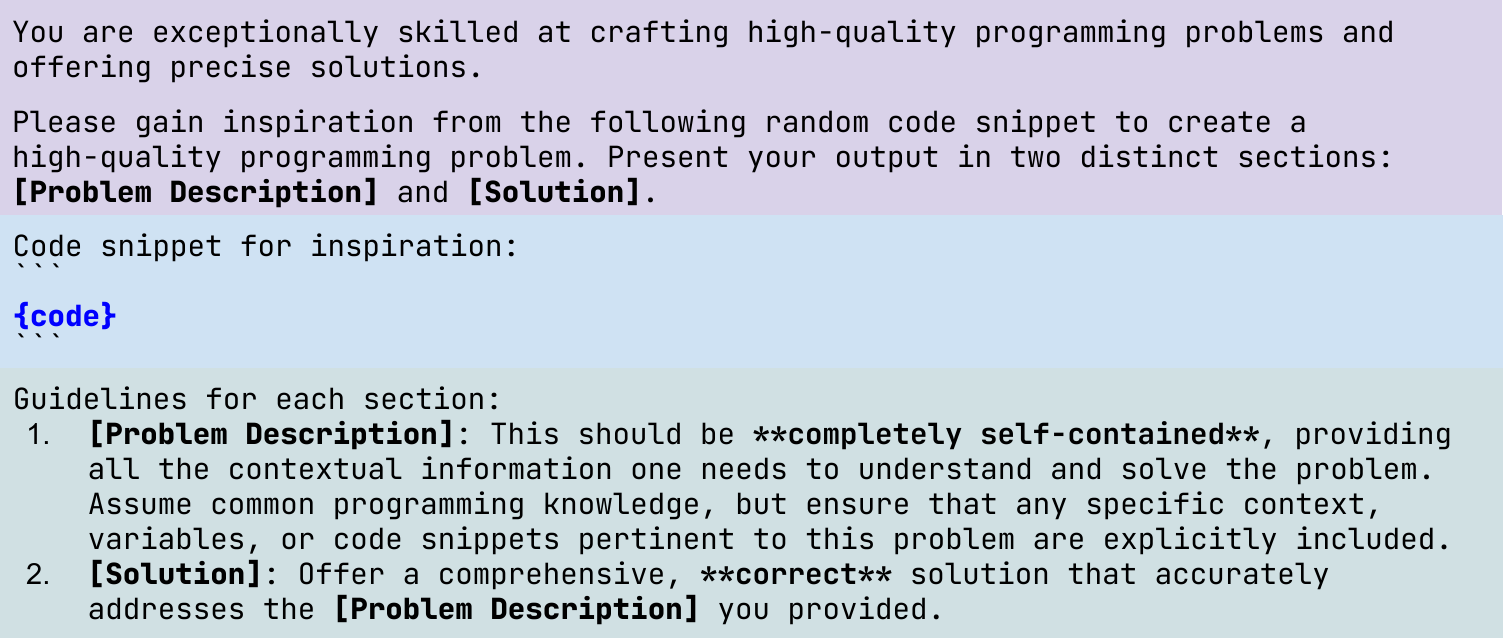}
\caption{The detailed prompt design for \tech}
\label{fig:prompt}
\end{figure*}
\Cref{fig:prompt} illustrates the prompt template of \tech{}, where the first section presents a high-level description of the task, the second section incorporates the code snippet, and the third section offers guidelines on the response.

\subsection{Qualitative Examples}
\label{apd:examples}
\begin{figure*}
\centering
\includegraphics[height=0.97\textheight]{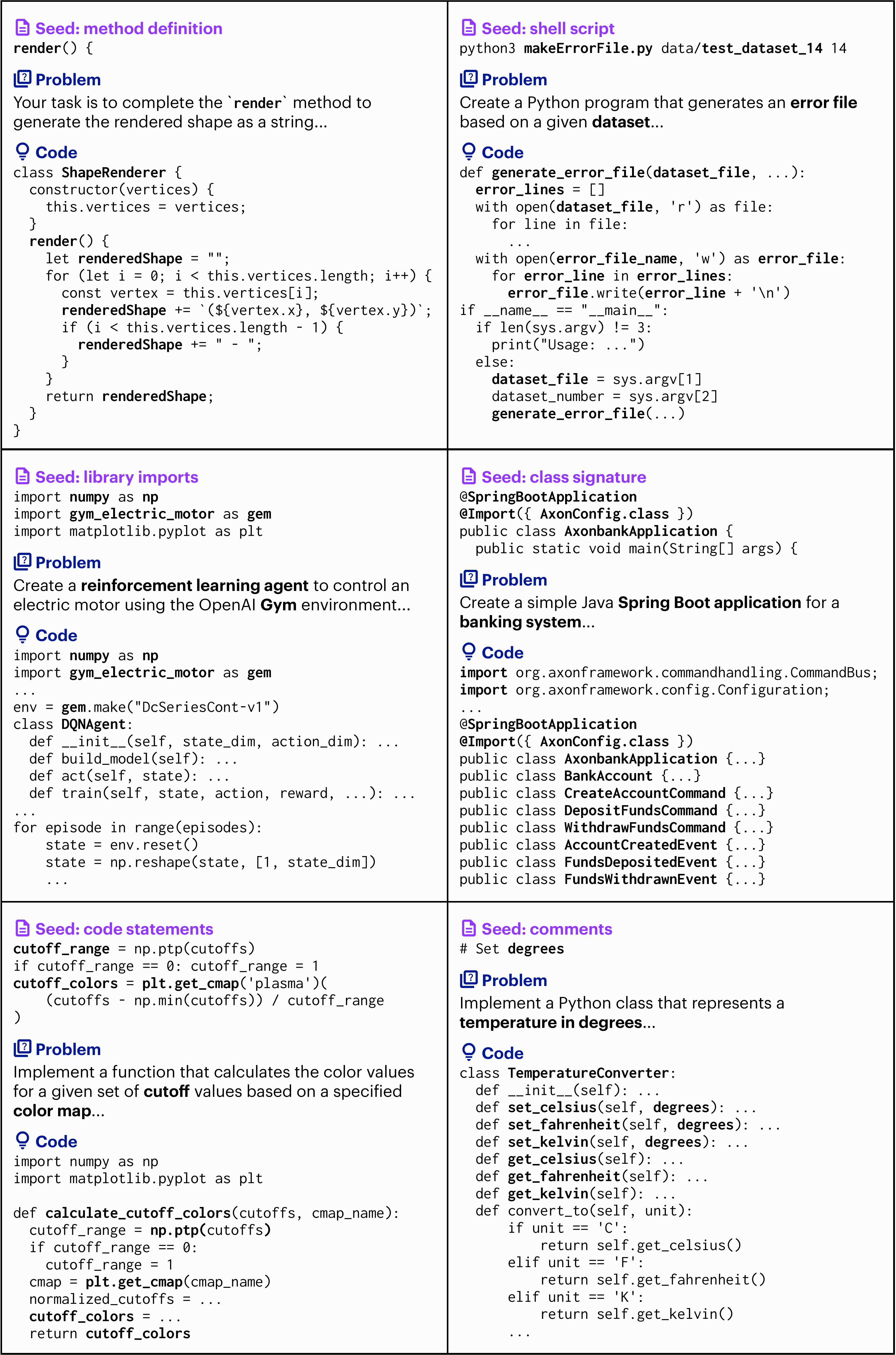}
\caption{More examples showing how \tech{} generates problems and solutions from seed code snippets. Detailed problem requirements, implementations, and explanations are omitted for brevity.}
\label{fig:full-examples}
\end{figure*}

\Cref{fig:full-examples} extends \Cref{fig:examples} and shows more qualitative examples of \tech{} generation, including outputs that are inspired by method definitions, shell scripts, library imports, class signatures, code statements, and code comments.

\subsection{Breakdown of \tech{} Dataset}
\label{apd:dataset}

\paragraph{Categories}
\begin{figure}
\centering
\includegraphics[width=\linewidth]{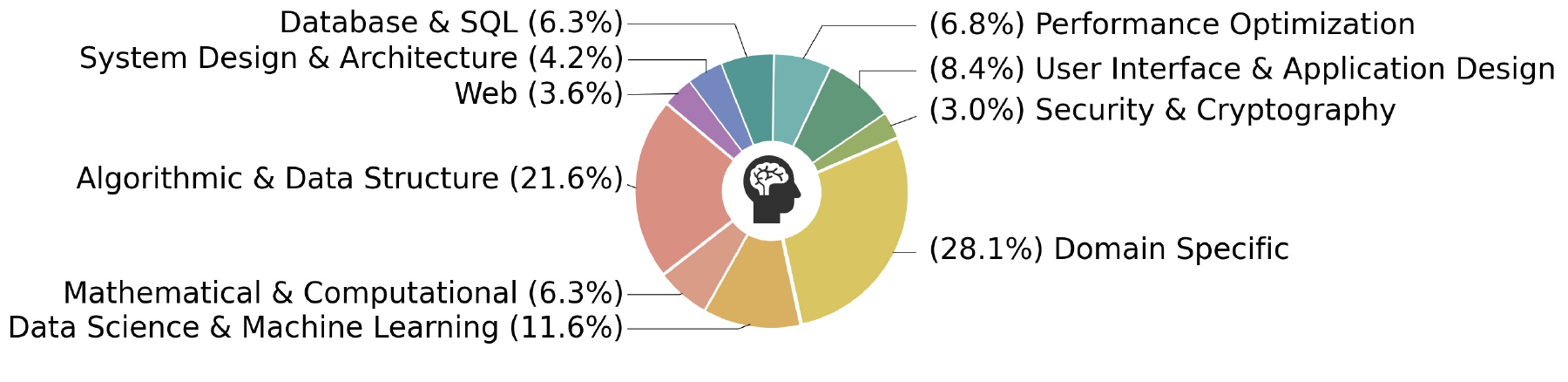}
\caption{The category constitution of \tech{}}
\label{fig:topic}
\end{figure}

To study the categories of \tech{}-generated data, we use \instructor~\cite{instructor}, which is one of the SOTA embedding models and can generate different text embeddings according to a task instruction. Inspired by \octopack~\cite{octopack} and the topic tags on \github, we manually designed 10 categories specific to coding. 
As shown in \Cref{fig:topic}, we calculate the cosine similarity between the embeddings of each sample in \tech{} and the 10 categories to obtain the category breakdown.
Overall, \tech{} exhibits diversity and balance across different categories.

\paragraph{Length distribution}
\begin{figure}
\centering
\includegraphics[width=0.5\linewidth]{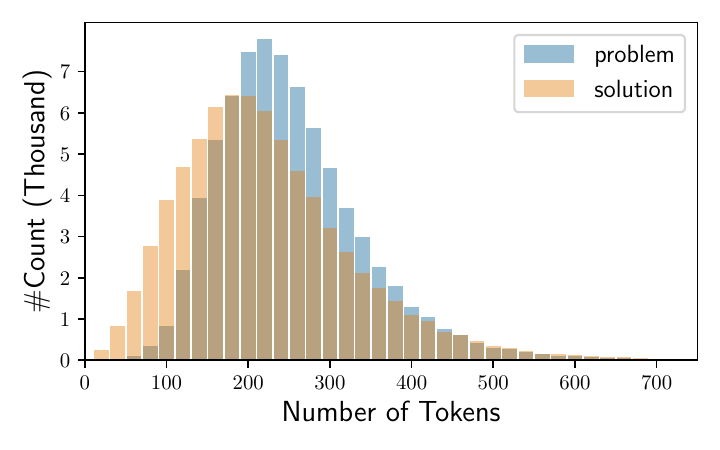}
\caption{Token count distribution of \tech{}-generated problems and solutions}
\label{fig:length}
\end{figure}

We depict the length distribution for both generated problems and solutions in \Cref{fig:length}. The x-axis represents the number of tokens in each problem/solution, while the y-axis shows the correspondent number of samples.

\section{Implementation Details}
\label{apd:setup}

\subsection{Data Generation}
We use \datagenmodel{} as the foundation model to do \tech{} due to its high cost-effectiveness.
We randomly extract 1--15 lines from each selected code document from \seedcorpus{} and let \datagenmodel{} imagine a self-contained coding problem and a correct solution. Given the numerous seed code snippets, we perform greedy decoding to maximize the consistency between the generated problems and solutions.

\subsection{Data Decontamination}
We apply data decontamination before training our \model{} and \modelx{} models.
Following \citet{starcoder}, we decontaminate both our 75K \tech{} dataset and the \evolcode{}~\cite{evolcode} dataset, an open-source reproduction of \evolinstruct{} generated by \gptfour~\cite{gpt4}, by removing exact matches from \humaneval{}~\cite{codex}, \mbpp{}~\cite{synthesisllm}, \dsonek{}~\cite{ds1000}, and \gsm{}~\cite{gsm8k}.
Eventually, we filtered out 9 problems for \tech{} dataset and 89 for \evolcode{}.

\subsection{Training}
We employ \codellamapy-7B and \dscoderbase~6.7B as the base \llm{s}.
To obtain \model{} series, we first finetune the base models on about 75K synthetic data generated through \tech{} using the \texttt{transformers} library from \hugging{}~\cite{huggingface}.
We finetune the base models for 2 epochs using two NVIDIA A100-80GB GPUs through the Distributed Data Parallel (DDP) module from PyTorch.
We set the initial learning rate at 5e-5 with 15 warmup steps and a linear scheduler. We use \adafactor{}~\cite{adafactor} as our optimizer and choose a batch size of 512 with a sequence truncation length of 1216.
To obtain \modelx{}, we continue to finetune \model{} models with the \evolcode{} dataset, an open-source \evolinstruct{} implementation containing about 110K samples.
We use the same hyperparameters except for 15 warmup steps and a 1024 maximum sequence length.

\section{More Evaluation Results}
\label{apd:evalmore}
\subsection{Evaluation on \apps{} for Competitive Programming}
We additionally evaluate \model{} on \apps{}~\cite{apps}, a benchmark suite of competitive programming problems. Following \citet{isselfrepair}, we select a subset of 300 problems from the \apps{} test set.
From \Cref{tab:apps}, we can observe that the \codellamapy{}-based \modelc{} significantly outperforms the base model and \wizardcoder{-CL}. \modelxc{-7B} is even better than \wizardcoder{-SC-15B} despite having less than half the number of parameters. Meanwhile, \dscoder{}-based \modelxd{} achieves the best result among all the evaluated baselines, substantially outperforming the instruction-tuned \dscoder{-6.7B-Instruct}.

\begin{table*}[htbp]
\centering
\caption{\Passat{1} results on \apps{} evaluated using greedy decoding in a zero-shot setting.}
\label{tab:apps}
\begin{booktabs}{
    colspec={@{}lrrrr@{}},
}
\toprule
Model & Introductory (60) & Interview (180) & Competition (60) & Overall (300) \\
\midrule
\wizardcoder{}-SC-15B & 21.7 & 6.1 & 1.7 & 8.3 \\
\codellamapy{}-7B & 3.3 & 2.8 & 0.0 & 2.3 \\
\wizardcoder{}-CL-7B & 10.0 & 3.9 & 1.7 & 4.7 \\
\modelc{}-7B & 18.3 & 5.6 & 1.7 & 7.3 \\
\modelxc{}-7B & 23.3 & 6.1 & 1.7 & 8.7 \\
\dscoder{}-6.7B-Base & 16.7 & 7.2 & 0.0 & 7.7 \\
\dscoder{}-6.7B-Instruct & 23.3 & 9.4 & 0.0 & 10.3 \\
\seprule
\modeld{}-6.7B & 20.0 & 8.9 & 1.7 & 9.7 \\
\modelxd{}-6.7B & 28.3 & 11.7 & 3.3 & \textbf{13.3} \\
\bottomrule
\end{booktabs}
\end{table*}

\subsection{Fill-in-the-Middle Evaluation on \dsonek{}}
\Cref{tab:ds1000-fim} shows the evaluation results of \modeld{} and \modelxd{} on \dsonek{}~\cite{ds1000} (Insertion format), assessing a model's fill-in-the-middle capability. In this experiment, we use \dscoder{} as the base model and exclude \codellamapy{}-based results, as \codellamapy{} does not support the fill-in-the-middle format. The results highlight Magicoder's superior performance in fill-in-the-middle tasks compared to all other evaluated baselines. This outstanding capability suggests that \model{} can serve as a valuable copilot for developers.

\begin{table*}[htbp]
\centering
\caption{\Passat{1} results on \dsonek{} (Insertion format) with $\temperature=0.2$, $\topp=0.5$, $\maxLen=1024$, and $\nsamples=40$.}
\label{tab:ds1000-fim}
\begin{booktabs}{
    colspec={@{}lrrrrrrr@{}},
}
\toprule
Model & NumPy & Pandas & PyTorch & SciPy & Sklearn & TensorFlow & Overall \\
\midrule
\wizardcoder{-SC}-15B & 35.1 & 20.4 & 30.4 & 28.9 & 32.3 & 37.8 & 28.6 \\
\dscoder{-6.7B-Base} & 36.3 & 28.6 & 15.8 & 19.3 & 32.8 & 35.1 & 29.3 \\
\dscoder{-6.7B-Instruct} & 44.1 & 27.3 & 38.2 & 30.8 & 38.4 & 29.6 & 34.6 \\
\seprule
\modeld{}-6.7B & 39.7 & 31.2 & 27.4 & 23.7 & 44.6 & 30.2 & 33.9 \\
\modelxd{}-6.7B & 43.3 & 29.5 & 39.2 & 26.2 & 44.5 & 36.2 & \textbf{35.9} \\
\bottomrule
\end{booktabs}
\end{table*}

\subsection{Impact of Removing Noisy Data}
\label{apd:noise}
In \Cref{sec:tech:decontaminate}, we highlight the benefits of preserving certain types of noise in instruction-tuning data. To support this argument, we removed data samples with partially implemented code, such as those containing \texttt{pass} or \texttt{NotImplemented} tokens in Python problems.
\Cref{tab:noise} shows the \passat{1} performance of \codellamapy{-7B} on \humanevalp{} when finetuned for 2 epochs on the raw \tech{} dataset compared to the same dataset with noisy samples removed. We can observe that retaining the noisy samples results in slightly better performance.

\begin{table*}
\caption{\Passat{1} on \humanevalp{} when finetuning \codellamapy{-7B} with and without noisy samples.}
\label{tab:noise}
\centering
\begin{booktabs}{
    colspec={@{}lrr@{}},
}
\toprule
Setting & Dataset Size & \humanevalp{} \\
\midrule
With noisy data & 75K & \textbf{55.5} \\
Without noisy data & 68K & 54.9 \\
\bottomrule
\end{booktabs}%
\end{table*}

\section{Cases where \tech{} Improves Code Understanding}
\label{apd:case}
\mdfsetup{innerleftmargin=5pt, innerrightmargin=5pt, frametitlefont=\rmfamily, frametitlebelowskip=0pt, nobreak=true}
\definecolor{backcolour}{rgb}{0.95,0.95,0.95}
\lstset{
  basicstyle={\small\ttfamily},
  columns=fullflexible,
  backgroundcolor=\color{backcolour},
  breaklines=true,
  frame=tb,
  framexleftmargin=2pt,
  framexrightmargin=2pt,
  framextopmargin=2pt,
  framexbottommargin=2pt,
  framerule=0pt,
    xleftmargin=2pt,
    xrightmargin=2pt,
    captionpos=b,
    belowcaptionskip=-5pt,
}

In this section, we compare the base model, \dscoder-6.7B-Base, and its \tech{}-tuned version, \model-DS-6.7B, in both in-domain and out-of-domain scenarios and demonstrate how \tech{} helps improve the base model's instruction following capability. We provide the base model with an appropriate prompt prefix to make sure it can produce a plausible output format.

\subsection{More Precise Understanding of Requirements}
The following example shows that after instruction tuning, the model can understand the requirement ``Check whether the input is valid'' more precisely. \dscoder-6.7B-Base only considers the integer case, while \model-DS-6.7B effectively checks if the parameter is of the correct type.
\begin{mdframed}[frametitle={Prompt: Write a function that takes an integer a and returns True if this integer is a cube of some integer number. Note: you should check whether the input is valid.}]
\begin{minipage}[t]{0.49\textwidth}
\begin{lstlisting}[language=Python, title=\dscoder-6.7B-Base]
def is_cube(a):
    if a <= 0:
        return False
    x = int(a ** (1/3))
    return x**3 == a
\end{lstlisting}
\end{minipage}
\hfill
\begin{minipage}[t]{0.49\textwidth}
\begin{lstlisting}[language=Python, backgroundcolor=\color{lightb}, title=\model-DS-6.7B]
def is_cube(a):
    if not isinstance(a, int) or a < 0:
        return False  # Invalid input

    # Check if the number is a perfect cube
    x = int(a ** (1/3))
    return x**3 == a
\end{lstlisting}
\end{minipage}
\end{mdframed}

\subsection{Handling Corner Cases}
The example below indicates that \model-DS-6.7B can correctly handle the corner case, where the input string has less than or equal to 2 characters, even if this is not explicitly stated in the prompt.
This is also an out-of-domain example since the Scala programming language is not included in \tech{} dataset.
\begin{mdframed}[frametitle={Prompt: Implement a Scala function that takes a string as input and returns a new string where all the characters are reversed except the first and last characters. Reply only with the function.}]
\begin{minipage}[t]{0.49\textwidth}
\begin{lstlisting}[language=Scala, title=\dscoder-6.7B-Base]
def reverseExceptFirstLast(str: String): String = {
  val chars = str.toCharArray
  val first = chars(0)
  val last = chars(chars.length - 1)
  val middle = chars.slice(1, chars.length - 1).reverse
  first + middle + last
}
\end{lstlisting}
\end{minipage}
\hfill
\begin{minipage}[t]{0.49\textwidth}
\begin{lstlisting}[language=Python, backgroundcolor=\color{lightb}, title=\model-DS-6.7B]
def reverseExceptFirstLast(input: String): String = {
  if (input.length <= 2) {
    input
  } else {
    val reversedMiddle = input.substring(1, input.length - 1).reverse
    input.charAt(0) + reversedMiddle + input.charAt(input.length - 1)
  }
}
\end{lstlisting}
\end{minipage}
\end{mdframed}

\subsection{Generalizing to Unseen Tasks}
We also show that after instruction tuning, \model-DS-6.7B can respond plausibly to unseen requests during training. This example asks the model to create an easy Python coding exercise, which the base model cannot interpret. Such kinds of tasks are also not incorporated in the \tech{} dataset.
\begin{mdframed}[frametitle={Prompt: Create an easy Python coding exercise}]
\begin{minipage}[t]{0.49\textwidth}
\begin{lstlisting}[language=Python, title=\dscoder-6.7B-Base]
def add_numbers(a, b):
    return a + b
\end{lstlisting}
\end{minipage}
\hfill
\begin{minipage}[t]{0.49\textwidth}
\begin{lstlisting}[language=Python, backgroundcolor=\color{lightb}, title=\model-DS-6.7B]
# Given a list of integers, return the sum of all the positive integers in the list.

def sum_positive_integers(nums):
    return sum(num for num in nums if num > 0)

# Test the function
numbers = [1, -2, 3, -4, 5, -6, 7, -8, 9, -10]
print(sum_positive_integers(numbers))  # Output: 25
\end{lstlisting}
\end{minipage}
\end{mdframed}

\section{Generating Complex Programs}
\label{apd:complex}
\lstset{backgroundcolor={}}
In this section, we present the cases where the best performing \modelx-DS-6.7B is capable of generating real-world, complex programs.

\subsection{Implementing Snake Game in Python}
{\vspace{5pt}}
\begin{mdframed}[frametitle={Prompt: Write a snake game in Python using the turtle library (the game is created by Magicoder).}, nobreak=false]

Response: Here is a simple snake game in Python using the turtle library. This game is a basic version and does not include many features that a full-fledged snake game would have.
\begin{lstlisting}[language=Python, basicstyle={\scriptsize\ttfamily}]
import turtle
import time
import random

delay = 0.1

# Score
score = 0
high_score = 0

# Set up the screen
wn = turtle.Screen()
wn.title("Snake Game by Magicoder")
wn.bgcolor("green")
wn.setup(width=600, height=600)
wn.tracer(0)  # Turns off the screen updates

# Snake head
head = turtle.Turtle()
head.speed(0)
head.shape("square")
head.color("black")
head.penup()
head.goto(0, 0)
head.direction = "stop"

# Snake food
food = turtle.Turtle()
food.speed(0)
food.shape("circle")
food.color("red")
food.penup()
food.goto(0, 100)

segments = []

# Pen
pen = turtle.Turtle()
pen.speed(0)
pen.shape("square")
pen.color("white")
pen.penup()
pen.hideturtle()
pen.goto(0, 260)
pen.write("Score: 0  High Score: 0", align="center", font=("Courier", 24, "normal"))


# Functions
def go_up():
    if head.direction!= "down":
        head.direction = "up"


def go_down():
    if head.direction!= "up":
        head.direction = "down"


def go_left():
    if head.direction!= "right":
        head.direction = "left"


def go_right():
    if head.direction!= "left":
        head.direction = "right"


def move():
    if head.direction == "up":
        y = head.ycor()
        head.sety(y + 20)

    if head.direction == "down":
        y = head.ycor()
        head.sety(y - 20)

    if head.direction == "left":
        x = head.xcor()
        head.setx(x - 20)

    if head.direction == "right":
        x = head.xcor()
        head.setx(x + 20)


# Keyboard bindings
wn.listen()
wn.onkeypress(go_up, "w")
wn.onkeypress(go_down, "s")
wn.onkeypress(go_left, "a")
wn.onkeypress(go_right, "d")

# Main game loop
while True:
    wn.update()

    # Check for a collision with the border
    if head.xcor() > 290 or head.xcor() < -290 or head.ycor() > 290 or head.ycor() < -290:
        time.sleep(1)
        head.goto(0, 0)
        head.direction = "stop"

        # Hide the segments
        for segment in segments:
            segment.goto(1000, 1000)

        # Clear the segments list
        segments.clear()

        # Reset the score
        score = 0

        # Reset the delay
        delay = 0.1

        # Update the score display
        pen.clear()
        pen.write("Score: {}  High Score: {}".format(score, high_score), align="center", font=("Courier", 24, "normal"))

    # Check for a collision with the food
    if head.distance(food) < 20:
        # Move the food to a random spot
        x = random.randint(-280, 280)
        y = random.randint(-280, 280)
        food.goto(x, y)

        # Add a segment
        new_segment = turtle.Turtle()
        new_segment.speed(0)
        new_segment.shape("square")
        new_segment.color("grey")
        new_segment.penup()
        segments.append(new_segment)

        # Shorten the delay
        delay -= 0.001

        # Increase the score
        score += 10

        if score > high_score:
            high_score = score

        pen.clear()
        pen.write("Score: {}  High Score: {}".format(score, high_score), align="center", font=("Courier", 24, "normal"))

    # Move the end segments first in reverse order
    for index in range(len(segments) - 1, 0, -1):
        x = segments[index - 1].xcor()
        y = segments[index - 1].ycor()
        segments[index].goto(x, y)

    # Move segment 0 to where the head is
    if len(segments) > 0:
        x = head.xcor()
        y = head.ycor()
        segments[0].goto(x, y)

    move()

    # Check for head collision with the body segments
    for segment in segments:
        if segment.distance(head) < 20:
            time.sleep(1)
            head.goto(0, 0)
            head.direction = "stop"

            # Hide the segments
            for segment in segments:
                segment.goto(1000, 1000)

            # Clear the segments list
            segments.clear()

            # Reset the score
            score = 0

            # Reset the delay
            delay = 0.1

            # Update the score display
            pen.clear()
            pen.write("Score: {}  High Score: {}".format(score, high_score), align="center", font=("Courier", 24, "normal"))

    time.sleep(delay)

wn.mainloop()
\end{lstlisting}
This game is a simple version of the classic snake game. The snake starts with a single segment and grows in size when it eats food. The game ends when the snake hits the border or its own body. The score is increased by 10 points for each piece of food eaten. The game speed increases as the snake grows.

\end{mdframed}

\subsection{Building Othello Game in Java}
{\vspace{5pt}}
\begin{mdframed}[frametitle={Prompt: Build a console-based Othello game in Java with row and column numbers shown on the board. The game should end when there are no more valid moves for either player.}, nobreak=false]

Response: Here is a simple console-based Othello game in Java. It does not include the graphical interface, but it does include the row and column numbers on the board.
\begin{lstlisting}[language=Java, basicstyle={\scriptsize\ttfamily}]
import java.util.Scanner;

public class Othello {
    private static final int BOARD_SIZE = 8;
    private static final char EMPTY = '-';
    private static final char BLACK = 'B';
    private static final char WHITE = 'W';
    private static final char[][] board = new char[BOARD_SIZE][BOARD_SIZE];
    private static char currentPlayer = BLACK;

    public static void main(String[] args) {
        initializeBoard();
        printBoard();

        while (true) {
            System.out.println("Player " + currentPlayer + ", enter your move (row column):");
            Scanner scanner = new Scanner(System.in);
            int row = scanner.nextInt();
            int col = scanner.nextInt();

            if (isValidMove(row, col)) {
                makeMove(row, col);
                printBoard();
                if (!hasValidMove()) {
                    System.out.println("Player " + currentPlayer + " has no valid moves. Game over!");
                    break;
                }
                currentPlayer = (currentPlayer == BLACK)? WHITE : BLACK;
            } else {
                System.out.println("Invalid move. Try again.");
            }
        }
    }

    private static void initializeBoard() {
        for (int i = 0; i < BOARD_SIZE; i++) {
            for (int j = 0; j < BOARD_SIZE; j++) {
                board[i][j] = EMPTY;
            }
        }
        board[3][3] = board[4][4] = BLACK;
        board[3][4] = board[4][3] = WHITE;
    }

    private static void printBoard() {
        System.out.print("  ");
        for (int i = 0; i < BOARD_SIZE; i++) {
            System.out.print(i + " ");
        }
        System.out.println();
        for (int i = 0; i < BOARD_SIZE; i++) {
            System.out.print(i + " ");
            for (int j = 0; j < BOARD_SIZE; j++) {
                System.out.print(board[i][j] + " ");
            }
            System.out.println();
        }
    }

    private static boolean isValidMove(int row, int col) {
        if (board[row][col]!= EMPTY) {
            return false;
        }
        for (int i = -1; i <= 1; i++) {
            for (int j = -1; j <= 1; j++) {
                if (i!= 0 || j!= 0) {
                    if (isValidDirection(row, col, i, j)) {
                        return true;
                    }
                }
            }
        }
        return false;
    }

    private static boolean isValidDirection(int row, int col, int rowDir, int colDir) {
        int newRow = row + rowDir;
        int newCol = col + colDir;
        while (newRow >= 0 && newRow < BOARD_SIZE && newCol >= 0 && newCol < BOARD_SIZE) {
            if (board[newRow][newCol] == EMPTY) {
                return false;
            }
            if (board[newRow][newCol] == currentPlayer) {
                return true;
            }
            newRow += rowDir;
            newCol += colDir;
        }
        return false;
    }

    private static void makeMove(int row, int col) {
        board[row][col] = currentPlayer;
        for (int i = -1; i <= 1; i++) {
            for (int j = -1; j <= 1; j++) {
                if (i!= 0 || j!= 0) {
                    flipDirection(row, col, i, j);
                }
            }
        }
    }

    private static void flipDirection(int row, int col, int rowDir, int colDir) {
        int newRow = row + rowDir;
        int newCol = col + colDir;
        while (newRow >= 0 && newRow < BOARD_SIZE && newCol >= 0 && newCol < BOARD_SIZE) {
            if (board[newRow][newCol] == EMPTY) {
                return;
            }
            if (board[newRow][newCol] == currentPlayer) {
                while (true) {
                    newRow -= rowDir;
                    newCol -= colDir;
                    if (newRow == row && newCol == col) {
                        break;
                    }
                    board[newRow][newCol] = currentPlayer;
                }
                return;
            }
            newRow += rowDir;
            newCol += colDir;
        }
    }

    private static boolean hasValidMove() {
        for (int i = 0; i < BOARD_SIZE; i++) {
            for (int j = 0; j < BOARD_SIZE; j++) {
                if (isValidMove(i, j)) {
                    return true;
                }
            }
        }
        return false;
    }
}
\end{lstlisting}
This program starts with a 8x8 board with two pieces in the center of the board, one black and one white. The player who goes first is black. The game ends when there are no more valid moves for either player. The player enters their move as two numbers, the row and the column. If the move is valid, the game makes the move and switches the player. If the move is not valid, the player is asked to enter a valid move.

\end{mdframed}

\subsection{Writing Specific Gradio Application}

{\vspace{5pt}}
\begin{mdframed}[frametitle={Prompt: Write a gradio (3.48.0) application for the following use case: Take an input image and return a 45 degree clockwise rotated image. You should also add text description under the output showing the rotation degree.}, nobreak=false]

Response: Here is a simple snake game in Python using the turtle library. This game is a basic version and does not include many features that a full-fledged snake game would have.
\begin{lstlisting}[language=Python,  basicstyle={\scriptsize\ttfamily}]
import gradio as gr
import numpy as np
from PIL import Image

def rotate_image(img):
    img = Image.fromarray(img)
    rotated_img = img.rotate(-45)
    return np.array(rotated_img), "45 degree clockwise rotation"

demo = gr.Interface(
    fn=rotate_image,
    inputs=gr.inputs.Image(type="numpy"),
    outputs=[gr.outputs.Image(type="numpy"), "text"],
    title="Image Rotation",
    description="Upload an image and get a 45 degree clockwise rotated image",
)

demo.launch()
\end{lstlisting}
This script defines a function \lstinline|rotate_image| that takes an image as input, rotates it 45 degrees clockwise using the PIL library, and returns the rotated image and a text description. The \lstinline|gr.Interface| function is then used to create a Gradio interface for this function.

The \lstinline|gr.inputs.Image(type="numpy")| line specifies that the input to the function should be an image, and the \lstinline|gr.outputs.Image(type="numpy")| line specifies that the output should be an image. The "\lstinline|text|" output type is used to display the text description.

Finally, \lstinline|demo.launch()| is used to launch the Gradio interface.

\end{mdframed}

\subsection{Developing Machine Learning Pipeline with PyTorch}

{\vspace{5pt}}
\begin{mdframed}[frametitle={Prompt: Build a simple neural network in Python using Pytorch to classify handwritten digits from the MNIST dataset. You should use CNN as the model structure, train the model for 5 epochs, draw a chart of the training loss, and show the final result.}, nobreak=false]

Response: Here is a simple implementation of a Convolutional Neural Network (CNN) using PyTorch to classify handwritten digits from the MNIST dataset.
\begin{lstlisting}[language=Python,  basicstyle={\scriptsize\ttfamily}]
import torch
import torch.nn as nn
import torchvision
import torchvision.transforms as transforms
import matplotlib.pyplot as plt

# Device configuration
device = torch.device('cuda' if torch.cuda.is_available() else 'cpu')

# Hyper-parameters 
num_epochs = 5
batch_size = 100
learning_rate = 0.001

# MNIST dataset 
train_dataset = torchvision.datasets.MNIST(root='../../data', 
                                           train=True, 
                                           transform=transforms.ToTensor(),  
                                           download=True)

test_dataset = torchvision.datasets.MNIST(root='../../data', 
                                          train=False, 
                                          transform=transforms.ToTensor())

# Data loader
train_loader = torch.utils.data.DataLoader(dataset=train_dataset, 
                                           batch_size=batch_size, 
                                           shuffle=True)

test_loader = torch.utils.data.DataLoader(dataset=test_dataset, 
                                          batch_size=batch_size, 
                                          shuffle=False)

# Convolutional neural network (two convolutional layers)
class ConvNet(nn.Module):
    def __init__(self):
        super(ConvNet, self).__init__()
        self.layer1 = nn.Sequential(
            nn.Conv2d(1, 16, kernel_size=5, stride=1, padding=2),
            nn.BatchNorm2d(16),
            nn.ReLU(),
            nn.MaxPool2d(kernel_size=2, stride=2))
        self.layer2 = nn.Sequential(
            nn.Conv2d(16, 32, kernel_size=5, stride=1, padding=2),
            nn.BatchNorm2d(32),
            nn.ReLU(),
            nn.MaxPool2d(kernel_size=2, stride=2))
        self.fc = nn.Linear(7*7*32, 10)
        
    def forward(self, x):
        out = self.layer1(x)
        out = self.layer2(out)
        out = out.reshape(out.size(0), -1)
        out = self.fc(out)
        return out

model = ConvNet().to(device)

# Loss and optimizer
criterion = nn.CrossEntropyLoss()
optimizer = torch.optim.Adam(model.parameters(), lr=learning_rate)

# Train the model
total_step = len(train_loader)
loss_list = []
for epoch in range(num_epochs):
    for i, (images, labels) in enumerate(train_loader):
        images = images.to(device)
        labels = labels.to(device)
        
        # Forward pass
        outputs = model(images)
        loss = criterion(outputs, labels)
        
        # Backward and optimize
        optimizer.zero_grad()
        loss.backward()
        optimizer.step()
        
        if (i+1) %
            print ('Epoch [{}/{}], Step [{}/{}], Loss: {:.4f}' 
                  .format(epoch+1, num_epochs, i+1, total_step, loss.item()))
            loss_list.append(loss.item())

# Test the model
model.eval()  # eval mode (batchnorm uses moving mean/variance instead of mini-batch mean/variance)
with torch.no_grad():
    correct = 0
    total = 0
    for images, labels in test_loader:
        images = images.to(device)
        labels = labels.to(device)
        outputs = model(images)
        _, predicted = torch.max(outputs.data, 1)
        total += labels.size(0)
        correct += (predicted == labels).sum().item()

    print('Test Accuracy of the model on the 10000 test images: {} %

# Draw a chart of the training loss
plt.plot(loss_list)
plt.title('Training Loss')
plt.xlabel('Epoch')
plt.ylabel('Loss')
plt.show()
\end{lstlisting}
This script first loads the MNIST dataset and then defines a Convolutional Neural Network (CNN) with two convolutional layers. The CNN is trained for 5 epochs using the Adam optimizer and the Cross Entropy Loss function. After training, the model is tested on the test dataset and the accuracy is printed. Finally, a chart of the training loss is drawn.
\end{mdframed}

\section{Limitations}
While \model{}s prove to be effective by being instruction-tuned with \tech{} and \evolinstruct{} generated data, 
we apply our techniques to models with no more than 7B parameters due to resource constraints.
This limitation hinders our ability to observe the method's impact on larger models. Moreover, the \tech{} prompt is sophisticated and may require a relatively strong model to comprehend the correct intent. Weaker models may fail to produce plausible instruction data.
Future research may explore applying \tech{} to models at different capacities in the context of self-training.

\end{document}